\documentclass[acmlarge, nonacm]{acmart}

\AtBeginDocument{%
  }

\copyrightyear{2026}
\acmYear{2026}
\setcopyright{cc}
\setcctype{by}
\acmConference[FAccT '26]{The 2026 ACM Conference on Fairness, Accountability, and Transparency}{June 25--28, 2026}{Montreal, QC, Canada}
\acmBooktitle{The 2026 ACM Conference on Fairness, Accountability, and Transparency (FAccT '26), June 25--28, 2026, Montreal, QC, Canada}
\acmDOI{10.1145/3805689.3806478}
\acmISBN{979-8-4007-2596-8/2026/06}

\usepackage{mathtools, amsfonts, bbold}
\usepackage{dsfont, mathrsfs} 

\DeclareMathAlphabet{\mathcal}{OMS}{cmsy}{m}{n} 

\newcommand{\XX}{\mathcal{X}}
\newcommand{\YY}{\mathcal{Y}}

\newtheorem{theorem}{Theorem}

\newtheorem{lemma}[theorem]{Lemma}

\theoremstyle{definition}

\theoremstyle{remark}

\makeatletter
\theoremstyle{acmplain}
\newtheorem*{rep@theorem}{\rep@title}
\newcommand{\newreptheorem}[2]{%
\newenvironment{rep#1}[1]{%
 \def\rep@title{#2 \ref{##1}}%
 \begin{rep@theorem}}%
 {\end{rep@theorem}}}
\makeatother

\newreptheorem{theorem}{Theorem}
\newreptheorem{lemma}{Lemma}

\begin{document}

\title[Costs of Pretending There Are Data-Generating Distributions]{The Costs of Pretending That There Are Data-Generating Probability Distributions in the Social World}


\author{Benedikt Höltgen}
\affiliation{%
  \institution{Hasso Plattner Institute, Universität Potsdam}
  \city{Potsdam}
  \country{Germany}}
\email{benedikt.hoeltgen@hpi.de}

\author{Robert C. Williamson}
\affiliation{%
  \institution{Eberhard Karls Universität Tübingen and Tübingen AI Center}
  \city{Tübingen}
  \country{Germany}}



\begin{abstract}
    Machine Learning research, including work promoting fair or equitable algorithms, often relies on the concept of a data-generating probability distribution. The standard presumption is that since data points are `sampled from' such a distribution, one can learn from observed data about this distribution and, thus, predict future data points which are also drawn from it. We argue, however, that such true probability distributions do not exist and that the rhetoric around them is harmful in social settings. We show that alternative frameworks focusing directly on relevant populations rather than abstract distributions are available and leave classical learning theory almost unchanged. Furthermore, we argue that the assumption of true probabilities or data-generating distributions can be misleading and obscure both the choices made and the goals pursued in machine learning practice. Based on these considerations, we suggest avoiding the assumption of data-generating probability distributions in the social world.
\end{abstract}

%
%

\ccsdesc[500]{Theory of computation~Machine learning theory}
\ccsdesc[300]{Computing methodologies~Supervised learning}
\ccsdesc[300]{Computing methodologies~Philosophical/theoretical foundations of artificial intelligence}

\keywords{probability distributions, data, fairness, i.i.d., society, multi-calibration}


\maketitle


\section{Introduction}
\label{s:intro}

We have grown so accustomed to probabilistic statements that we hardly question what they entail. Particularly in Machine Learning (ML) and Statistics, probabilities and probability distributions are ubiquitous.
Students are required to learn the basics of probability theory since virtually all theoretical work in ML is embedded in a probabilistic framework.
This makes sense since most of ML is concerned with prediction, and predictions contain uncertainty---which can often be captured through probability theory.
However, probabilities play a central role not only as predictions but also as true distributions that our data is sampled from; this is the general theoretical framework that we use to think about machine learning, for example, to derive error guarantees \citep{vapnik1982}.
But while we can do a lot of maths with probability theory, we still lack a reliable understanding of when and why our probabilistic models and frameworks actually work in practice.\footnote{This can serve as an illustration of John von Neumann's point that `if people do not believe that mathematics is simple, it is only because they do not realize how complicated life is' \citep{alt1972}.}
The purpose of this paper is to argue that, at least in social settings, we should avoid the true data-generating distribution (Gen-D) framework that is ubiquitous in Machine Learning as well as Statistics.
This applies particularly to work promoting fair or equitable algorithms.

It is a very common assumption in all areas of ML to say that an observed set of data points is sampled from some true joint distribution.
This allows us to mathematically capture the (more or less stable) patterns we see in the world, and to formalise learning.
This is a particular choice of a mathematical model; to avoid confusion with the specific `models' trained on concrete data (such as neural networks), we will call it the Gen-D \textit{framework} here.
As researchers, we often make statements like `we assume that our data points $\{(x_i,y_i)\}_{i=1}^N$ are sampled independently and identically distributed (i.i.d.) from the true distribution over $\XX \times \YY$', even though we know that this is an idealisation.
One may seek consolation in saying that all models are wrong anyway---but it is important to know in which ways they are wrong, in order to anticipate when they fail.
The still common (since easy to work with) i.i.d. assumption is increasingly criticized in the community;
however, the suggested adjustments typically only address either of the i's, especially the `identical', via frameworks for distribution shift or data corruption.
In contrast, we already see a more fundamental problem with saying that our data is somehow sampled from some abstract true distribution.\footnote{A line of work that does not rely on this assumption is conformal prediction \citep{shafer2008}, although its reliance on exchangeability assumptions makes it vulnerable to similar critiques.}
The Gen-D assumption often also motivates the suggestion that the goal of learning is to approximate this distribution in terms of the `Bayes-optimal predictor' $P(Y|X)$.
And for particularly stable settings, such as gambling devices or pseudo-random number generators, data-generating distributions can indeed be a useful idealisation.
In this paper, however, we argue that in social settings, when making predictions about people, we should not assume the existence of a true distribution, even as an idealisation.

We start, in Section~\ref{s:no_true_distr}, by arguing that true generative distributions do not exist, even if, within the limits of stable problem settings, they can be useful modelling tools.
In Section~\ref{s:not_needed}, we then debunk their perceived raison d'\^{e}tre, namely that they are necessary to model the stabilities which we see in practice and, in particular, learning and generalisation; we show that one can modify classical learning theory so that it works with finite populations.
In Section~\ref{s:dangerous}, we show how the Gen-D framework obscures how Machine Learning works and what it can provide.
In particular, it may suggest that model-building is approximation rather than construction, as it was originally conceived.
This has important implications, for example, on the interpretation of theoretical trade-offs between fairness and accuracy \cite{hardt2016,corbett2017,menon2018}, and their relation to empirical observations about model multiplicity and the possibility of less discriminative algorithms \cite{cooper2024,black2024,holtgen2025ftu}.
We also argue that the Gen-D framework conceals the modelling choice of selecting an input space due to its framing of the learning problem, which can be an obstacle both to further research and to open discussion in applications.
In Section~\ref{s:objective}, we highlight a sometimes underappreciated implication of our foregoing discussions, namely, that ML models are far from objective, as they depend on many design choices.
Section~\ref{s:conclusion} concludes.


\section{Gen-Ds do not really exist}
\label{s:no_true_distr}

In this section, we argue that there are no true probabilities and, by implication, no true generative distributions in ML.
Still, people tend to have particularly strong intuitions in favour of objective probabilities in gambling settings such as coin flips or dice throws: there do seem to be `correct' probabilities for such games of chance---but this is misleading.  
As put by Michael Strevens, probabilities in gambling settings `have attained a certain kind of stability under the impact of additional information. This stability gives them the appearance of objectivity, hence of reality, hence of physicality’ \citep[p. 31]{strevens2006}.
Indeed, this stability is the very reason why games like roulette are used for gambling because there is no easy way to beat the casino.\footnote{Though see \citep{thorp1998} for an anecdote of how Edward Thorp and Claude Shannon used hand-made devices to beat casinos at roulette.\label{fn:thorp}}

As Quantum Mechanics (QM) is often thought to prove the existence of true probabilities, we feel compelled to briefly dispute this.
First, QM is consistent with determinism, e.g. via Bohmian mechanics \citep{durr2009}.
Second, even if the world were indeterministic, this could mean that the QM-probability of one and the same coin turning up heads is above 50\% in one situation and below 50\% in another situation, due to minuscule differences in background conditions.
But as we neither have access to the complete world state nor have the computational resources to compute the whole dynamics, the (non-)existence of such QM-probabilities is actually irrelevant for the interpretation of the macro-scale probabilities we deal with in everyday life and in virtually all settings where ML is applied.
As Nancy Cartwright observes, the natural sciences and especially physics have focused on finding latent quantities that stand in stable relationships to each other, whereas in social settings we typically consider quantities that are of direct interest or easy to measure.
And, as she puts it, `to suppose that there really is some probability measure over [such quantities], you need a lot of good arguments’  \citep[p. 325]{cartwright1999}.
In sum, at least in social settings, it seems fair to say that probabilities are always constructed rather than discovered \citep{holtgen2024, spiegelhalter2024}.



But how about the data-generating distributions that are typically assumed in ML?
One approach to try to justify the assumption of such a distribution could be to interpret it as the empirical distribution that \textit{would} be observed if the whole population were to be queried.
Take the example of predicting whether people in the US with a certain set of attributes will commit a crime within the next year.
Consider the (in practice unobserved) empirical distribution that captures the whole population of the US for, say, 10 or 20 years.
One could then regard an actually collected dataset (of, say, thousands of data points) as a sample from that empirical distribution (constructed from millions of data points).
The two problems we will now point out are particularly clear for very fine-grained input spaces $\XX$, that is, when we collect many attributes.

\begin{figure}[t]
    \centering
    \includegraphics[width=.49\linewidth]{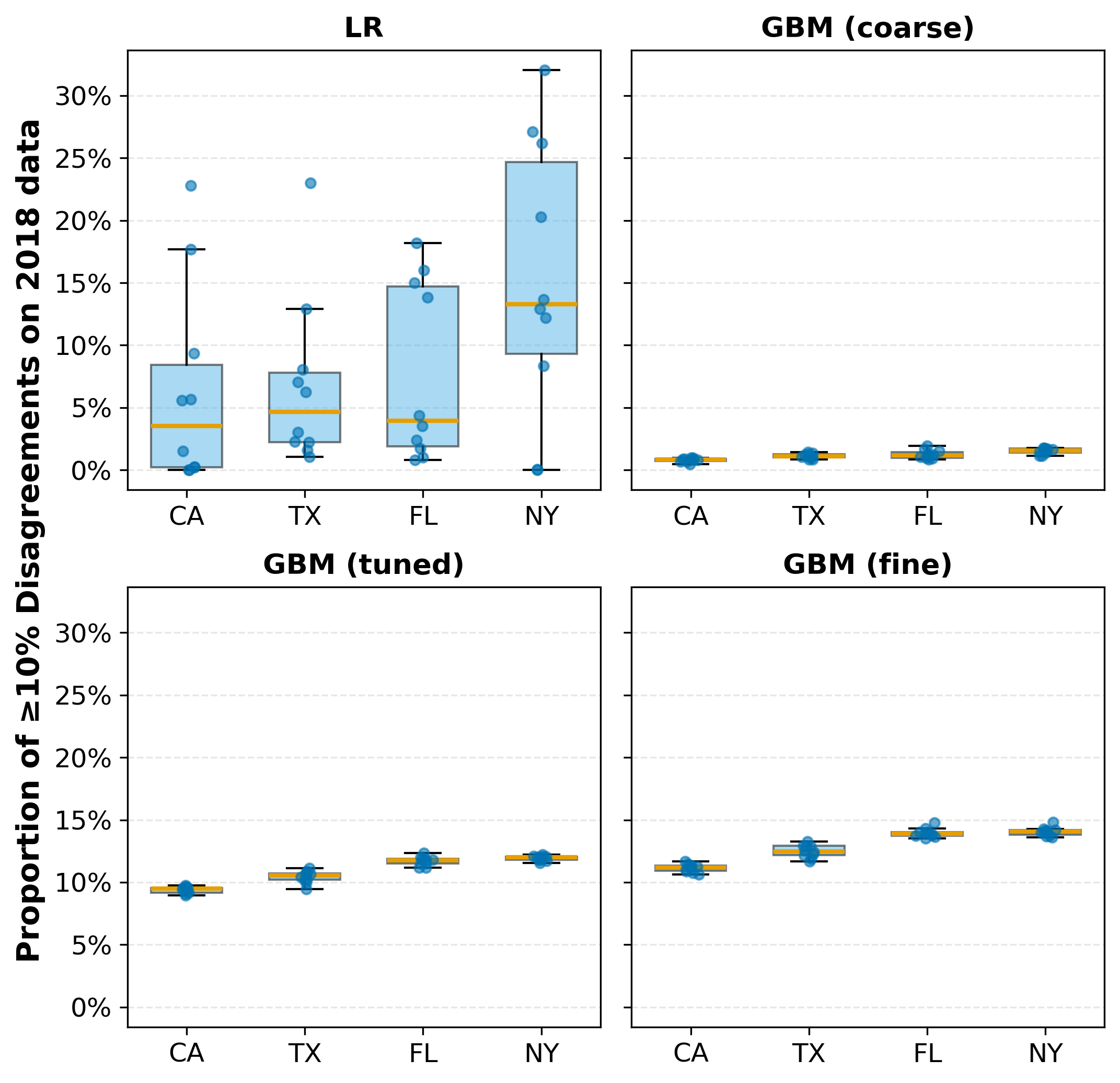}
    \hfill
    \includegraphics[width=.49\linewidth]{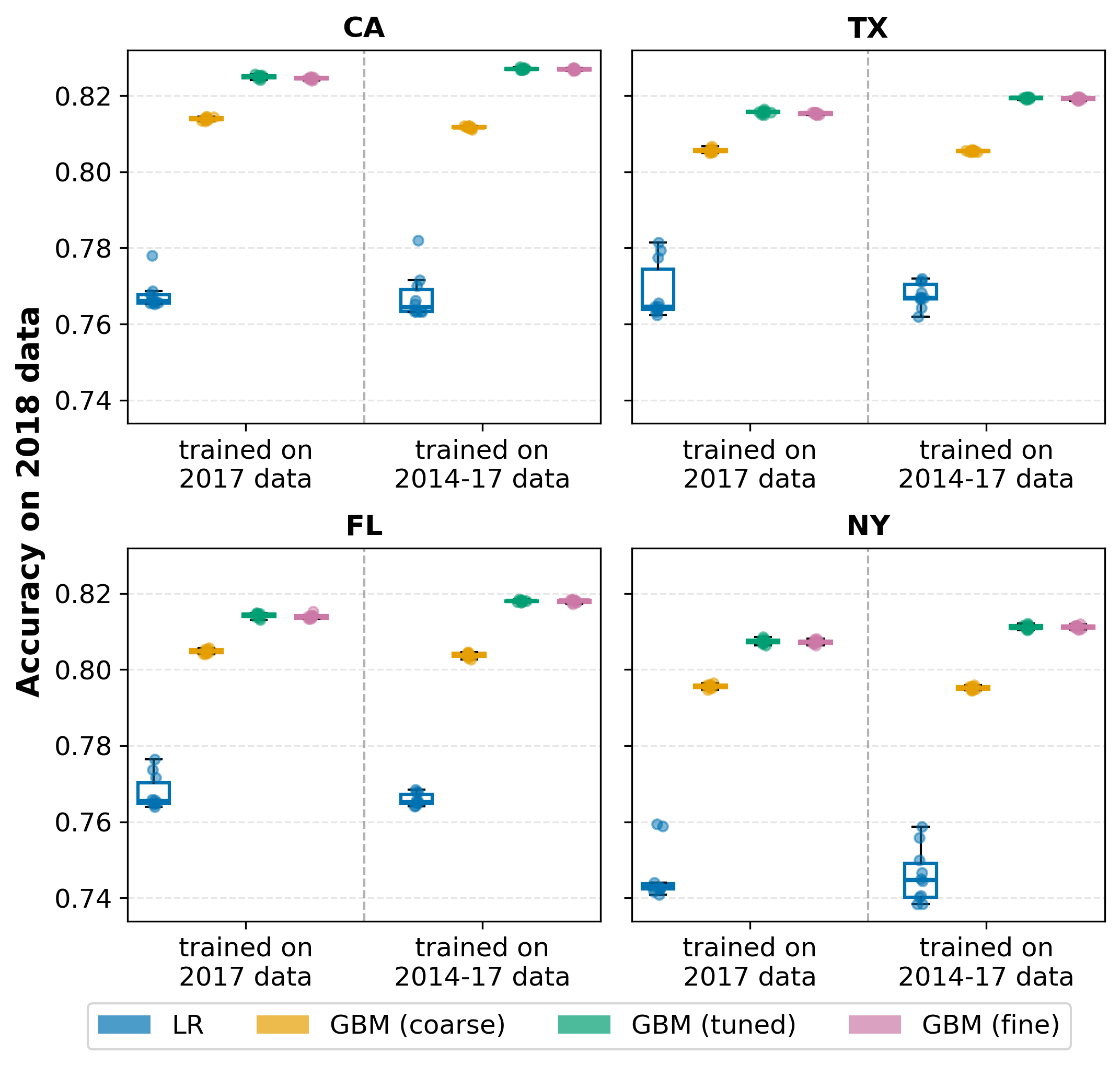}
    \caption{Model multiplicity based on choice of training population in \texttt{ACS Income}, for different model classes on the four largest US states. 
    Left panel: Each dot corresponds to a pair of models, one trained on 2017 data and one on 2014-2017 data (both using the same subset of the 2017 data). Coarse GBMs use default hyperparameters while fine GBMs use those considered optimal on the whole US dataset in previous work \citep{cruz2024,holtgen2025ftu}, see Appendix~\ref{app:exp_details}. For the tuned GBMs, around 10\% of data points in 2018 have at least 10\% disagreement in the prediction, depending on whether the models were trained only on the last year or on the last four years. Coarser GBMs show less multiplicity, finer GBMs show more; LR results are quite~volatile.\linebreak[1]
    Right panel: Accuracy on 2018 data when trained on either the previous year or the previous four years. Tuned GBMs significantly outperform GBMs with default settings as well as LR models and have similar performance to the finer GBMs. For all models and states, accuracy does not depend much on the chosen training population.}
    \label{fig:multiplicity}
\end{figure}

One problem is that the `true' distribution depends very strongly on our implicit choices (beyond the choice of abstraction that we will discuss in Section~\ref{ss:choice}).
In particular, it depends on what we take to be the population described by the empirical distribution.
For example, if there are few data points per $x \in \XX$, then the true conditional probability $P(Y|X)$ depends a lot on the time frame that the distribution is supposed to cover: 
if there are only a few people with the characteristics $x$, then $P(Y|X=x)$ can strongly depend on whether we take into account 10 or 20 years, given that the observed ratios are typically not stable then.
In the ML context, this can be seen, for example, in degradation of performance over time.
In the example of predicting income based on US census data, the accuracy of models trained in one year continuously degrades every year \cite{ding2021}, see also our results on individual states in Appendix~\ref{app:degradation} and other results across domains by \citet{vela2022}.
We now want do draw attention to another aspect of this, also considering \texttt{ACS Income} \citep{ding2021}:
The predictions even of our best models, which are often considered close to Bayes-optimal as classifiers, significantly depend on the choice of population.
In Figure~\ref{fig:multiplicity}, we show that around 10\% of the predictions of tuned models change by at least 10\%, depending on whether we consider the last year or the last four years
These numbers apply to the four largest US states with very large datasets and are even higher for datasets of smaller states.
The proportions are lower for some suboptimal coarser models, but even higher for comparably accurate but finer models (more details in Appendix~\ref{app:exp_details}).
If the distribution depends on the year, how about the day or second a person is `sampled'?
Beyond time considerations, `transfer from one state to another gives unpredictable results in terms of predictive accuracy and fairness criteria.' \citep[p. 9]{ding2021}
In general, making the choice of the considered population explicit is important for applying ML in social settings.
This suggests a different understanding of generalisation, to which we will come back later.


Now take the example of predicting credit default, where there is likely \emph{e.g.} a general trend that connects higher income with lower rates of credit default.
Assume, then, for some $x_1 \in \XX$ that includes very high income as well as otherwise favourable attributes, there are only three data points among the millions which fall into this description; assume further that two of these three people happened to default for whatever reasons.
Would we want to say the `true probability' of default for that combination of attributes is really $P(Y=1|X=x_1)=2/3$?
Now the problem is not just that we \textit{cannot} learn such a volatile conditional distribution, but that we do not \textit{want} to:
We aim to learn general patterns in the data that `generalise',\footnote{The common, unspecific talk of `generalisation' leaves open its scope, i.e. the population to which it should generalise. As argued above, this typically goes unnoticed precisely because learning is always discussed in the context of a true distribution with undefined scope. If we want to predict well on unseen data, however, we need some reason to believe that it is similar to our training or validation data---but in what sense? This needs to be specified with reference to the model we use and the metrics we care about; as Nelson \citet[p. 18]{Goodman1972} already suspected, `rather than similarity providing any guidelines for inductive practice, inductive practice may provide the basis for some canons of similarity'.\label{fn:generalise}} not ratios between data points that will change dramatically when one or two more data points are observed.
Take the mentioned credit example.
Assume that we have access to a look-up table (or oracle) that tells us all the conditional probabilities.
Is it sensible to deny people with attributes $x_1$ a loan because, despite their favourable profile, the default ratio among three data points is $2/3$?
Given that both the data seen in training and the data seen at deployment are thought to be sampled from the empirical distribution, the considered population must cover both the past and the future.\footnote{We might want to say that they come from different distributions between which a distribution shift occurs---but this distribution shift can be almost arbitrary if each conditional probability depends only on two or three data points. More generally, the distribution shift approach makes comparisons between populations even less tangible by critically relying on the relationship between two abstract entities.}
So some of the three points may be in the past, and some may be in the future.
Therefore, we cannot even argue that predicting the true empirical distribution helps us to optimise our decisions in the future:
The Gen-D framework fails to account for the phenomenon that data points seen in the past will not be seen again in the future: in practice, we sample \emph{without} replacement (for the relevance of this, see also footnote~\ref{fn:roth} below).
A central problem with the conventional framework is, then, that it takes the data to originate from distributions; actually, it is the other way around.


\section{Gen-Ds are not needed to understand learning}
\label{s:not_needed}

Now in spite of these theoretical considerations, the Gen-D framework seems to provide a useful framework for modelling and understanding learning.
In the following, we demonstrate that the assumption of data-generating distributions is not needed for this.
Empirically, we do see that ML models actually do learn patterns, even on human behaviour---so if we do not learn some empirical distribution, then must there not be an abstract true distribution after all? 
This conclusion is particularly intuitive because the Gen-D framework seems to be the only framework at our disposal to model such stability, via the law of large numbers.
We now demonstrate that it is an unwarranted conclusion, both by drawing a historical parallel and by introducing an alternative framework.

With only a little stretching, we can trace precursors to this kind of argument back to the early statisticians in the 19th century like Adolphe Quetelet who, observing, \emph{e.g.}, a `frightening regularity with which the same crimes are reproduced’, tried to find the laws that govern society \citep{porter1986}. 
This also led to a widespread belief in true probabilities which should govern human behaviour according to general laws \citep{hacking1990}, very similar to the true data-generating distributions referred to in ML.
To explain the stable rates of crimes within countries, Quetelet came up with an abstract construct called the average man (`l'homme moyen'), which, `defined in terms of the average of all human attributes in a given country, could be treated as the ``type'' of the nation, the representative of a society in social science comparable to the center of gravity in physics' \citep{porter1986}.
The notion of a true conditional distribution in ML today is different basically only in the level of granularity: we now have more data such that we can exploit apparent stabilities on a finer level.
For then as for now, however, we can find alternative frameworks that capture such stabilities.

One idea is that if we cut a large enough sequence of data points with binary labels (`crime', `no crime') in half (two consecutive years), then for most partitions, the label ratios in both halves will be roughly similar.
This can also be framed as drawing from an urn without replacement, where we calculate the number of possible draws of $N$ out of $2N$ balls for which the ratio of `crime' balls is roughly the same as for the remaining $N$.
We capture this through the counting measure ($\mu$ in the results below), quantifying the number of admissible draws without any notion of probability. \textit{If} we assigned the same probability to each possible draw (a notion of exchangeability), we could interpret this measure as a probability---but this is an additional assumption, a special case, in this framework.
We can extend this to a model of learning, only slightly adapting the work of Vapnik.
The following result is similar to Theorem~A.2 in \cite[p. 170]{vapnik1982}; the difference is that we bound the two notions of generalisation error in finite populations rather than the error when generalising to a true distribution.
The proof relies on a modified Symmetrisation Lemma, of which we prove two versions; we chose a slightly simplified version of Vapnik's notation here to facilitate comparisons between results and proofs.

\begin{lemma}[Symmetrisation Lemma, version 1]
    \label{lem:symm1}
    \ \\
    Take an urn with $l+k$ balls, with $k \geq l > 2/\epsilon$.
    Let $u(\alpha)$ denote the error rate of $\alpha$ on the whole urn.
    Then
        \begin{align*}
            \mu &\left[ \sup_\alpha \left| u(\alpha) - v_{tr}(\alpha) \right|  > \epsilon \right] 
            < 2 \cdot \mu \left[ \sup_\alpha \left| v_{te}(\alpha) - v_{tr}(\alpha) \right| > \left(\frac{1}{2}+\frac{l}{k}\right) \epsilon \right].
        \end{align*}
\end{lemma}

Note that sampling without replacement converges to sampling with replacement for $k\to \infty$. 
Accordingly, $k\to \infty$ implies $\left(\frac{1}{2}+\frac{l}{k}\right) \epsilon \to \epsilon/2$ which gives Vapnik's original Symmetrisation Lemma (`The Basic Lemma',  p.168).

\begin{lemma}[Symmetrisation Lemma, version 2]
    \label{lem:symm2}
    \ \\
    Take an urn with $l+k$ balls, with $k \geq l > 2/\epsilon$.
    Let $u'(\alpha)$ denote the error rate of $\alpha$ on the remaining urn.
    Then
        \begin{align*}
            \mu \left[ \sup_\alpha \left| u'(\alpha) - v_{tr}(\alpha) \right|  > \epsilon \right]
            < 2 \cdot \mu \left[ \sup_\alpha \left| v_{te}(\alpha) - v_{tr}(\alpha) \right| > \epsilon / 2 \right].
        \end{align*}
\end{lemma}

For $k=l$ we have $u' = v_{te}$ and this version is weaker than the previous version's bound with $\epsilon' := \epsilon/2$. given that $2 \cdot |u-v_{tr}| = |v_{te}-v_{tr}|$, which then gives the RHS bound $3/4 \epsilon$.
This is because above we take the cumulative from $u - \epsilon'/2 = u' - \epsilon/2 - \epsilon/4$, whereas here we take it from $u' - \epsilon/2$.
Note that for $k \to \infty$, we get $u' \to u$ and, thus, Vapnik's original Symmetrisation lemma.
With these results under our belt, we can now prove the generalisation bound.

\begin{theorem}\label{LT_theorem}
    \ \\
    Take a set of $l+k$ data points in $\mathcal{X} \times \{0,1\}$ and draw a training set of size $l$ without replacement.
    Consider a hypothesis class $\Lambda$ of finite VC dimension $h$.
    Denote by $u(\alpha)$, $v_{tr}(\alpha)$, and $u'(\alpha)$ the error rate of $f_\alpha \in \Lambda$ on the whole set of $k+l$ points, on the training set of $l$ points, and on the remaining $k$ points, respectively.

    Then for $k \geq l > 2/\epsilon$ with $\epsilon >0$, we can bound the proportion of draws in which the generalisation error exceeds $\epsilon$ by
    \begin{align*} 
    \mu \left[ \sup_{\alpha \in \Lambda} \left| u(\alpha) - v_{tr}(\alpha) \right| > \epsilon \right]
    \ < \
    9 \frac{(2l)^h}{h!} \cdot \exp \left(-\epsilon^2l \left(\frac{1}{2}+\frac{l}{k}\right)^2\right) 
    \end{align*}
    and 
    \begin{align*} 
    \mu \left[ \sup_{\alpha \in \Lambda}\left| u'(\alpha) - v_{tr}(\alpha) \right| > \epsilon \right]
    \ < \ 
    9 \frac{(2l)^h}{h!} \cdot \exp \left(-\epsilon^2l/4\right),\quad\quad\quad\
    \end{align*} 
    where $\mu$ is the counting measure on possible draws of size $l$.
\end{theorem}

At first glance, the successful loss minimisation of ML models across many domains seems to strengthen the case for the Gen-D framework.
But as the above result shows, we can understand learning also without invoking a data-generating distribution---indeed, the ideas of pattern recognition and generalisation do not rely on such a distribution at all.
The theorem shows that for large enough sample sizes, the error on the sample will be similar to the error on the whole set for the vast majority of possible samples.
Note that this provides a more concrete notion of generalisation, featuring data that can (in principle) be seen in the future rather than metaphysical notions of expectations under theoretical distributions; indeed, this relates directly to the generalisation error between training and test set that is commonly investigated in practice.
At this point, the reader may ask, all models are wrong, so what? 
There is some alternative framework that was in some sense already implicit in the Gen-D framework, so what?
After all, science commonly works with idealisations \cite{potochnik2017}, acting as if the models were true \cite{appiah2017}.
In the remainder of this paper, we argue that there are good reasons to stop working with thinking in terms of data-generating distributions, at least in social settings.


\section{Gen-Ds facilitate misinterpretations}
\label{s:dangerous}



Drawing on both philosophical and theoretical considerations, we have argued that generative distributions are fictions that are not even that useful.
In the remainder, we argue that their invocation can even have negative effects in practice, as they can suggest an understanding of ML as an approximation of true probabilities, depending less strongly on particular choices of data collection and training.


\subsection{Gen-Ds obscure what is being optimised}
\label{ss:minim_loss}

The ML literature often suggests implicitly or explicitly that the goal of learning is to recover the data-generating distribution or to approximate the Bayes-optimal classifier, which is a function of the true conditional $P(Y|X)$.
This is indeed appealing if one believes that there is such a true distribution to recover, estimate, or approximate.
But, as we have discussed, there is actually no such distribution.
As put by a physicist, `the phrase ``estimating a probability'' is just as much a logical incongruity as  ``assigning a frequency''' \citep[p. 120]{jaynes1985}.
The goal of learning, as also formalised in empirical risk minimisation and statistical decision theory, is to predict well on average, to minimise a loss that reflects our preferences and needs (on a population of interest, see footnote~\ref{fn:generalise}).
Recall the example from Section~\ref{s:no_true_distr} concerning the prediction of credit default, where two out of three people with some $x$ in a population default: 
There is no true probability that we want to estimate; rather, we want to learn patterns or decision rules that work well on average (see also \citep{holtgen2024}).

This is not always appreciated nowadays, arguably due to the optimism associated with modern Deep Learning and Big Data.
This is reflected in the extremely popular textbook by central figures in this area, which states that `[i]deally, we would like to match the true data-generating distribution $p_{data}$, but we have no direct access to this distribution' \citep[p. 130]{goodfellow2016}.
In a recent real-world example, the Austrian public employment service rebuked criticism of their logistic regression model, developed to help allocate job training programmes, by claiming that its prediction `follows the real chances on the labour market as far as possible' (as cited in \cite{allhutter2020}).
Such statements, which appear to become more common also with the more recent optimism based on the ever-increasing availability of data and compute \citep{boyd2012}, mistake error minimisation for an approximation of true probabilities.
This has arguably also to do with the use of more complex model classes, as one can theoretically show that in the limit of infinite data, an idealisation related to Gen-D, models like (infinitely big) neural networks or random forests can approximate or will converge to any data-generating distribution. 

The role or meaning of probabilities is also relevant to the way we evaluate and select models in practice.
As observed by Cynthia Dwork, `without an answer to this definitional question, we don’t even know what it is that the ideal algorithm should satisfy’ \citep{dwork2022}.
The arguments against the Gen-D assumption, in particular, highlight the task dependence of ML-based predictors.
In the conventional framework, the conditional distribution $P(Y|X)$ provides the best predictor for every task:
It would `suffice' to approximate this $P(Y|X)$ and use it for every task that requires the prediction of $Y$, as this would minimise any proper loss.
Indeed, such arguments are often made.
However, if there is no true underlying distribution, it becomes difficult to separate the trained predictor from the specific learning tasks.
Indeed, much recent work has focused on getting multi-calibrated \cite{hebert2018} or omni-predictors \cite{gopalan2023}, deriving guarantees based on the assumption that we can draw ever more data from the same stable distribution with replacement.
The underlying intuition of a `march towards truth'  (attributed to Cynthia Dwork \cite{roth2025}) is clearly in conflict with a central insight of the present paper, that there is no such thing as a true data-generating distribution.
It also surfaces in the assumption that multi-calibration--type algorithms like RECONCILE \cite{roth2023} are thought to work `independent of how rich the feature space is', even up to individual people \cite{roth2025}---which clearly cannot guarantee ever-improving predictions about people with unique profiles that were not represented in the training dataset. 
The maths checks out, however, because the Gen-D framework assumes that we sample people \textit{with} replacement---which is at odds with assuming that observed events and predicted events have no overlap.
We contend that the Gen-D framework is inadequate for capturing the limitations of such approaches.\footnote{Recent work has highlighted that multi-calibration is also possible with hardly any assumption, by coupling it with defensive forecasting strategies for the online learning setting \citep{dwork2025,perdomo2025}. However, the calibration guarantees in this setting only guarantee predicting the base rate (per pre-defined group) while (swap) regret guarantees only serve as a comparison against very specific alternatives. They are all very different to the `march towards truth' that we criticise here and do not have the discriminative power of forward-looking ML algorithms. Clearly, without making some assumptions about the future, we cannot say how good our predictions will be in absolute terms.}

In practice, the observation that there is no correct predictor that is adequate for every task goes beyond the difficulty of different loss functions leading to different approximations (\emph{i.e.} models): now, there is no true probability that could serve as a mutual reference point for the different models.
It also goes beyond the empirical observation of the `Rashomon effect' or `predictive/model multiplicity', of which Leo Breimann already said in 2001 that `its effect on conclusions drawn from models needs serious attention' \citep[p. 206]{breiman2001}.
It is, however, only recently that the community is increasingly grappling with this phenomenon, that there are often many models which perform equally well on average for a given task while disagreeing significantly on individual predictions \cite{marx2020, black2022, holtgen2025ftu}.
Illustrations of this phenomenon on datasets for credit default, rearrests, and medical diagnoses in human populations were recently given by \citep{watson2023}; they show that discrepancies in individual predictions of $20-40$ percentage points are common for different models with near-optimal loss---with the \textit{same} model class, \textit{same} abstraction, and \textit{same} loss function.
In the Gen-D framework, this multiplicity is only an approximation problem, a practical nuisance due to a constrained model class or a small sample size; without this framework, we can see that predictive multiplicity is actually a fundamental conceptual issue (see also \citet{heljakka2022}).

This is also important for the notion of fairness-accuracy trade-offs \cite{hardt2016,corbett2017,menon2018}. 
These analyses are based on the Bayes-optimal classifier, that is, they assume a data-generating distribution. 
Given this framing, fairness considerations are necessarily seen as deviations from optimality.
Empirical Work, however, shows that it is often possible to get less discriminative models without reductions in performance metrics like average accuracy \cite{cooper2024, black2024, holtgen2025ftu}.
While the trade-off result could be interpreted as speaking purely about theoretical concepts, they are often thought to also apply to actual models in practice (otherwise, what would be their relevance?):
Even fairness researchers sometimes explicitly assert that such results should hold in practice, stating e.g. that considered models `are likely close to Bayes optimal under the squared loss' \citep[p. 2]{cruz2024} (echoing \citet{ding2021}).
As the trained models can rarely be considered close to the \textit{empirically} optimal model (otherwise, we would not need machine learning with its inductive biases), such discussions do implicitly or explicitly assume the existence of the data-generating distributions and, thus, of a correct model that is being approximated.
Indeed, recent theoretical results that take into account the larger difference between trained models and empirically optimal ones show that there is typically much room for less discriminative but equally accurate models \citep{holtgen2025ftu}.

Another interesting aspect relating to model performance is that belief in true probabilities may make us overestimate the benefits of individualistic interventions as compared to structural ones.
A recent extensive empirical investigation into a programme aimed at preventing students from dropping out of high school in Wisconsin using ML finds that the best way forward would not lie `in identifying students at risk of dropping out within specific schools, but rather in overcoming structural differences across different school districts' \citep{perdomo2025wisconsin}.
The former approach \textit{does} seem more powerful, though---if we believe that students have a true individual risk that is out there for us to find (see also \citep{perdomo2024, shirali2024}).


\subsection{Gen-Ds take abstractions for granted}
\label{ss:choice}

So far, we have been discussing the assumption of a true generative distribution for a given input and label space, in particular with the assumption of true conditional probabilities $P(Y|X=x)$.
In contrast to that, we now focus on what ML predictions are actually used for---that is, the individual events that we represent through abstractions $x \in \XX$.
This could, for example, be the probability that you in particular, rather than anyone with your attributes $x$, will commit a crime within the next year.
This relates to the so-called `reference class problem' for frequentism \citep{reichenbach1949}: if we take the probability of some event $A$ to consist in the ratio (or relative frequency) of how often events similar to $A$ do occur, then this will depend on the deployed notion of similarity.
An illustrative example is the risk of developing a disease in the next year, say lung cancer: the risk is usually taken to be the ratio of people `like you' getting the disease.
But this ratio will depend on whether we take into account age, gender, country of residence, whether you smoke, what you eat, et cetera.
If we take into account too many attributes, we will end up with a single data point, you.
This again shows that there is no single correct probability, that probabilities are inherently model-dependent.\footnote{In an example showing how the Gen-D framework can obscure this, recent work claims to make progress on a version of the reference class problem, yet assuming that we can sample (\emph{with} replacement) from a distribution that already defines (possibly trivial) true probabilities, thereby also \emph{fixing} a level of abstraction \citep{roth2023, roth2025}. As a more general point, the freedom to choose an adequate level of abstraction need not be seen as a `problem' to solve \citep{holtgen2024}. \label{fn:roth}}

In this sense, an ML model by itself never provides probabilities for individual events (or `follows real chances'); they always predict probabilities $P(Y=1|X=x)$ that are conditioned on a certain abstraction.
In practice, we automatically map individual events to their representation in the input space $\XX$, so this nuance is easily overlooked---especially as literature on ML typically takes the input space $\XX$ as given.
Consequently, the construction of abstractions, that is, the choice of the input space, is often not discussed enough in ML.\footnote{This arguably has to do with the observation that `everyone wants to do the model work, not the data work' \citep{sambasivan2021}.}
However, the choice of how to represent the events we want to predict determines how we carve up the world and, thus, which events are seen as similar.
And since learning from experience or data always depends on the notion of similarity (via inductive biases favouring smooth predictors in the case of ML), this, in turn, influences the resulting predictions.
This can be seen as a generalisation of the reference class problem to cases where we do not just predict the average label of all events that we represent through the same abstraction:
The outputs of more complicated prediction methods that are based on more flexible notions of similarity also depend on the mode of abstraction---which allows the modeller to chose a suitable abstraction for a given task \cite{holtgen2024}.

We have argued previously that it is already problematic to even assume true conditional probabilities $P(Y|X=x)$ for modelling.
It seems even more problematic to argue that some selection of attributes $x$ should be the `correct' representation of John Doe for predicting, \emph{e.g.}, whether he will commit a crime or find a job within the next year.
And even with a fixed abstraction, the predicted probabilities depend on other choices about the data, such as the considered time frame, et cetera.
So, again, there is no true probability to be found---but in contrast to the former, there can be a lot of disagreement on how to model such more complicated events and how to calculate probabilities.

\begin{figure}[h]
    \centering
    \includegraphics[width=\linewidth]{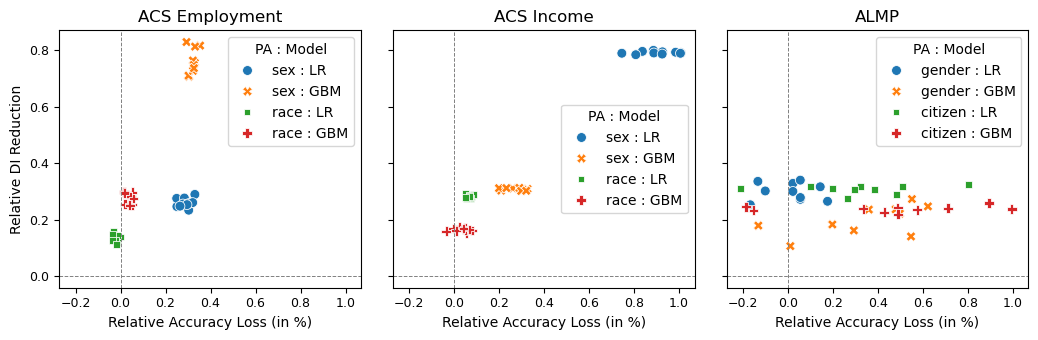}
    \caption{Reproduced from \citet{holtgen2025ftu}: Relative reduction in Accuracy vs in Disparate Impact (DI) across three large datasets, for logistic regression (LR) and gradient boosted trees (GBM). Each dot corresponds to a model pair with the same train/test split, one with and one without the protected attribute (PA). Omitting the PA reduces DI by $15$-$80\%$ (y-axis) but hardly reduces---and sometimes improves---the accuracy (x-axis).  The original publication \cite{holtgen2025ftu} contains additional details.}
    \label{fig:scatter_results}
\end{figure}
Given that it is quite clear that the choice of abstraction matters for the predictions, it is remarkable how little this choice is sometimes discussed (depending on who you talk to).
As a result, decision makers sometimes choose to include demographic attributes like gender in practice, but we still do not have a solid understanding of when this leads to more discrimination; and a lack of established best practices necessarily leads to a lack of accountability.
Recent work \citep{holtgen2025ftu} suggests that more such attributes do not necessarily lead to better predictions, but may exacerbate discrimination (see our Figure~\ref{fig:scatter_results})---contrary to the myth of Big Data \citep{boyd2012}, but also contrary to our idealised theoretical models.
We, thus, argue that the framing of ML problems in terms of a fixed space $\XX \times \YY$ with a true joint distribution has contributed to this:
Partly because it is outside this problem framing, and partly because the true distribution already conveys that whatever features one selects, one can approximate true probabilities.
It has been noted that because of the way ML is done, `the abstraction is taken as given and is rarely if ever interrogated for validity. This is despite the fact that abstraction choices often occur implicitly, as accidents of opportunity and access to data' \citep[p. 60]{selbst2019}.
The Gen-D framework, thus, induces us to underestimate the inferential leap in identifying a conditional probability $P(Y|x)$ to commit a crime for your attributes $x$ with `your' probability to commit a crime: your data was not sampled from a distribution, it was constructed as a profile of you.


\section{True probabilities and the illusion of objectivity}
\label{s:objective}

Algorithmic predictions are often presented as objective; in the previously mentioned example on job training allocations, predictions were said to aim for `the \textit{highest objectivity possible} in the sense that the predicted integration likelihood [\dots] follows the real chances on the labour market as far as possible' (as cited in \cite{allhutter2020}, emphasis added).
This perception of objectivity is partly an effect of quantification in general \citep{porter1995}, but partly also specific to ML applications, where the data are often thought to `speak for themselves' \citep{boyd2012, moss2021objective}.
As Justin Joque puts it, `statistics and algorithms effectively launder nonobjective forms of violence and bias, giving them greater stability, only to be fed back into these systems as the initial conditions for the next round' \citep[p. 174]{joque2022revolutionary}.\footnote{We are grateful to an anonymous reviewer for pointing us to this work.}

If there were true probabilities, then getting a low(er) average loss could be seen as an indication that one's predictions get close(r) to these true probabilities, vindicating the model irrespective of the choices that went into designing and training it.
The choices would only serve the purpose of getting close to the true probabilities, and with enough data, we may think that we get close enough to make the specific choices less relevant.
If, however, there is no such true probability to approximate, no such indirect vindication is possible; hence, the choices need to be justified more explicitly \citep{holtgen2025ftu}.
Rather than seeing models as constructed on top of available data, even critical work often repeats the phrasing (from a data science textbook, in this case) that statistical models are `estimating' individual values of interest such as `a crime location or offender' \citep[p. 423]{strikwerda2021} or `an individual’s level of risk' \citep[p. 382]{jacobs2021}.
Crucially, the idea that machine learning models approximate individual probabilities contributes to understating the relevance of specific modelling choices; as documented by \citet{moss2021objective}, this is `how applied machine learning researchers contribute to the idea that numbers can speak for themselves’ (p. 21).
Based on this, ‘the products of machine learning come to instead appear as separate from their labor […], not only automatic or magic, but also natural and inevitable’ \citep[p. 13]{moss2021objective}.
Indeed, this idea is directly reflected in the predominant theoretical understanding of machine learning.
This idea of approximating objective values is particularly dangerous for probabilistic predictions, as---in contrast to predicting observable attributes like height---we do not observe individual prediction errors.
By highlighting that probabilities in general are constructed rather than discovered, we contribute to previous work arguing that algorithmic predictions are not the objective arbiters of truth that they are sometimes taken to be.

Indeed, many `value-laden choices' \citep{allhutter2020} by ML engineers go into designing such systems without later being discussed as such.\footnote{This fits well into the older literature on values in science more generally \citep{rudner1953, douglas2000}. Also the choice of `ontology' (\emph{i.e.} categories, quantities to measure, ...) in science depends on the specific goals \citep{danks2015} and non-epistemic values \citep{ludwig2016}.}
Because the creators of such models often do not discuss their choices, letting the results `speak for themselves', `the products of machine learning come to instead appear as separate from their labor -- [...] not only automatic or magic, but also natural and inevitable' \citep[p. 13]{moss2021objective}.
This concerns not only the construction of models but also of data \citep{fletcher2025credal}; as argued in \citep{leonelli2019}, `there is no such thing as ``raw data'', since data are forged and processed through instruments, formats, algorithms, and settings that embody specific theoretical perspectives on the world'.
The deployment of Machine Learning systems for consequential decisions, thus, needs actual justification in relation to the data and modelling choices.
This means that these choices need to be more openly discussed and recorded---recent years have seen suggestions for standardised documentation (which are common in other areas of engineering) to record such choices as well as intended uses for models \citep{mitchell2019} and datasets \citep{gebru2021, hutchinson2021}.

It should also be noted that minimising average loss is, first and foremost, meant to be useful for the decision-maker and the population as a whole; it is harder to justify it from the individuals' perspective if it is not possible to claim that their `objective risk' is being estimated.
For example, the choice of abstraction is often more difficult to justify from the individual's perspective, as it determines with whom you are considered equivalent (and who is considered similar, via inductive biases).
We are, however, not arguing here categorically against making decisions about people based on data that was collected from other people.
After all, that is also something that human decision-making relies on.
What we want to caution against is a premature justification of decisions informed by ML-based systems based on the perceived objectivity of such systems, which is reinforced by a na\"ive view on probability.
We are not samples from distributions; we are the real thing.


\section{Conclusion}
\label{s:conclusion}

In this paper, we have problematised the assumption of true data-generating distributions as it often appears in the context of Machine Learning.
We emphasised that probabilities in general are always constructed rather than discovered \citep{holtgen2024}.
We have demonstrated that in the case of ML, the ubiquitous assumption of a true data-generating distribution (Gen-D) is not always a benign one, especially when people are involved.
In particular, it contributes to the air of objectivity that often circulates in predictions based on data science. 
As an alternative, we suggested a more parsimonious framework for understanding learning in social settings based on finite populations.
In addition to the epistemic modesty it conveys, it can also lead to new intuitions and insights.
While working with probability distributions is sometimes considered more elegant, this not only comes at the price of clouding the modelling assumptions\footnote{It has also been shown that in practice, statements made in the language of probability more easily mislead people than the same statements made in terms of relative frequencies in populations \citep[Chapter 9]{gigerenzer2008}.} but is also less expressive \citep{meng2022}.
In contrast, modelling finite populations directly requires empirically testable modelling assumptions, which can pave the way for more accountability in model development. 

Based on this work, we would like to see further research into three directions in particular.
One direction relates to the choice of abstraction/representation, that is, of the input space/covariates, as highlighted in Section~\ref{ss:choice}.
Although it is quite clear that such choices have a large impact on the resulting predictions, it seems that---for reasons we discussed---very little research so far has addressed this issue.
We, thus, encourage researchers as well as practitioners to follow Lily Hu's `call to arms' \citep{hu2023}, to think more about what she calls `variable construction' and we prefer to call the choice of abstraction.
Another direction is to further develop theory within the finite population framework; this will sometimes involve easy adaptations, as in Theorem~\ref{LT_theorem}, but open up more possibilities, e.g., by going beyond i.i.d. sampling assumptions via non-probability sampling as developed in the survey sampling literature \citep{meng2022}.

The third direction concerns the limits of our models:
While the standard story suggests that there are distributions waiting for us everywhere, not all tasks are created equal in practice.
We, thus, need more empirical insights into which domains and quantities are actually stable to admit fruitful predictions.
Indeed, some domains may lend themselves more readily to qualitative inquiry methodologies developed in the social sciences \citep{beuving2015}.
We will need to assess the stability of the systems against the requirements of our prediction models and tasks (footnote~\ref{fn:generalise}), to decide which models (if any) are suitable in which settings.
Many systems work remarkably well without us understanding why, but they also often break down unexpectedly, especially in the social world, sometimes with serious consequences for affected populations \citep{wang2022}.
Establishing best practices for stating concrete and domain-specific inductive assumptions can be one element in avoiding more such failures, but also pave the way for more accountability.
Assuming a stable distribution by default, and rhetoric like `approximating the current state' obscures the need for this.

The choice of framework is, in general, more relevant for model evaluation than model training: 
The evaluation depends directly on the framing of the problem, especially on how performance on current data is thought to inform future performance.
A departure from the Gen-D framing should not be understood as a departure from validation, but as a call to think more concretely about the relationships between training, validation, and deployment populations, rather than assuming they are sampled from the same distribution (or from distributions within an arbitrary $\epsilon$-ball on some abstract divergence measure).
We highlighted that the idea of `recovering the true data-generating distribution', is, in most cases, misguided---the aim is, rather, to find a model that performs well for some choice of loss function and aggregation scheme.
The goal of ML should be to find a good model, not the true model; this seems to have been clearer in earlier days when Machine Learning was more often understood as Pattern Recognition.
Note also that the aggregation scheme for losses need not---and, arguably, often should not---be the average, even though it is easy to work with and explicitly used in expected risk minimisation (and also in this paper).
In many settings, we are interested in other distributional properties that pay more attention to worst-case performance, such as the class of `fairness risk measures' \citep{williamson2019}.

The choice of training algorithm, on the other hand, appears to be more independent of the chosen framework and has largely eluded rigorous theoretical justification.
Deep Learning is not really motivated by any particular formal framework---indeed, seminal papers introducing e.g. Transformers \citep{vaswani2017} make no reference to data-generating distributions.\footnote{Note that our work primarily addresses contexts where data points represent people. Although language is also quintessentially social, not all of our arguments generalise to language modelling. While there are related pitfalls, such as the treatment of dialects, many languages may be sufficiently stable to make Gen-D assumptions benign.}
For the most part, training then depends on the problem formulation only indirectly, via the evaluation criterion: we choose training methods that work well for the chosen evaluation criterion.
Still, looking at the topic from a different angle can be a fresh breeze that provides insights into why and when such models work well.
Conversely, models that are built explicitly on `subjectivist' Bayesian statistics, such as Variational Autoencoders, are often still justified through theoretical results that employ Gen-D assumptions \citep{kingma2013}.
We would like to stress again that the purpose of the paper is not to advocate for a full-blown rejection of the Gen-D framework in general, as it seems adequate in less complex settings like gambling.

Nancy Cartwright argues that `extending our favourite theories to treat new problems of new kinds' is not dangerous only `when there is good empirical evidence for the promise of the proposed approach' and `not objectionable so long as we are clear about what we are doing and what it is going to cost us, and we are able to pay the price' \cite[p. 333]{cartwright1999}.
We have argued that the costs of pretending that there are data-generating distributions in social settings are high and that there are no real benefits (beyond convention/inertia).
When it comes to decision-making about people, there is too much risk of contributing to claims of objectivity and the evasion of a thorough justification of modelling choices.
The language we use and assumptions we make around ML systems are particularly relevant for the literature on fair and just deployment of algorithms; all too often, societal concerns are seen as departures from simply approximating the true probabilities.
We take issue with `simply', `approximating', and `the true probabilities'.


\section*{Generative AI Usage Statement}

Claude Sonnet 4.5 was used for literature searches, grammar verification, and coding assistance. No AI tools were used to generate text for the manuscript.



\begin{acks}
We would like to thank Rabanus Derr, Nina Effenberger, and Christian Fr\"ohlich for helpful feedback on earlier versions. 
Benedikt Höltgen worked on this project as a PhD student at the University of Tübingen.
This work was funded by the Deutsche Forschungsgemeinschaft (DFG, German Research Foundation) under Germany’s Excellence Strategy—EXC number 2064/1—Project number 390727645 as well as the German Federal Ministry of Education and Research (BMBF): Tübingen AI Center, FKZ: 01IS18039A. The authors thank the International Max Planck Research School for Intelligent Systems (IMPRS-IS) for supporting Benedikt H\"oltgen.
\end{acks}

\bibliographystyle{ACM-Reference-Format}
\bibliography{library_thesis}

@incollection{goodman1972,
  author    = {Goodman, Nelson},
  title     = {Seven strictures on similarity},
  booktitle = {Problems and Projects},
  publisher = {Bobbs-Merrill},
  year      = {1972},
  address   = {Indianapolis},
  pages     = {437--447}
}

@inproceedings{selbst2019,
  title={Fairness and abstraction in sociotechnical systems},
  author={Selbst, Andrew D and Boyd, Danah and Friedler, Sorelle A and Venkatasubramanian, Suresh and Vertesi, Janet},
  booktitle={Proceedings of the 2019 Conference on Fairness, Accountability, and Transparency},
  pages={59--68},
  year={2019}
}

@inproceedings{williamson2019,
  title={Fairness risk measures},
  author={Williamson, Robert and Menon, Aditya},
  booktitle={International Conference on Machine Learning},
  pages={6786--6797},
  year={2019},
  organization={PMLR}
}

@phdthesis{moss2021objective,
  title={The objective function: Science and society in the age of machine intelligence},
  author={Moss, Emanuel D},
  year={2021},
  school={City University of New York}
}

@article{allhutter2020,
  title={Algorithmic profiling of job seekers in {Austria}: How austerity politics are made effective},
  author={Allhutter, Doris and Cech, Florian and Fischer, Fabian and Grill, Gabriel and Mager, Astrid},
  journal={Frontiers in Big Data},
  pages={5},
  year={2020},
  publisher={Frontiers}
}

@inproceedings{black2022,
  title={Model multiplicity: Opportunities, concerns, and solutions},
  author={Black, Emily and Raghavan, Manish and Barocas, Solon},
  booktitle={Proceedings of the 2022 ACM Conference on Fairness, Accountability, and Transparency},
  pages={850--863},
  year={2022}
}

@article{black2024,
  title={Less discriminatory algorithms},
  author={Black, Emily and Koepke, John Logan and Kim, Pauline and Barocas, Solon and Hsu, Mingwei},
  journal={Georgetown Law Journal},
  volume={113},
  number={1},
  year={2024}
}

@article{boyd2012,
  title={Critical questions for big data: Provocations for a cultural, technological, and scholarly phenomenon},
  author={Boyd, Danah and Crawford, Kate},
  journal={Information, Communication \& Society},
  volume={15},
  number={5},
  pages={662--679},
  year={2012},
  publisher={Taylor \& Francis}
}

@article{breiman2001,
  title={Statistical modeling: The two cultures (with comments and a rejoinder by the author)},
  author={Breiman, Leo},
  journal={Statistical Science},
  volume={16},
  number={3},
  pages={199--231},
  year={2001},
  publisher={Institute of Mathematical Statistics}
}

@inproceedings{corbett2017,
  title={Algorithmic decision making and the cost of fairness},
  author={Corbett-Davies, Sam and Pierson, Emma and Feller, Avi and Goel, Sharad and Huq, Aziz},
  booktitle={Proceedings of the 23rd ACM SIGKDD International Conference on Knowledge Discovery and Data Mining},
  pages={797--806},
  year={2017}
}

@inproceedings{cruz2024,
  title={Unprocessing Seven Years of Algorithmic Fairness},
  author={Cruz, Andr{\'e} and Hardt, Moritz},
  booktitle={The Twelfth International Conference on Learning Representations},
  year={2024}
}

@article{danks2015,
  title={Goal-dependence in (scientific) ontology},
  author={Danks, David},
  journal={Synthese},
  volume={192},
  pages={3601--3616},
  year={2015},
  publisher={Springer}
}

@article{ding2021,
  title={Retiring adult: New datasets for fair machine learning},
  author={Ding, Frances and Hardt, Moritz and Miller, John and Schmidt, Ludwig},
  journal={Advances in Neural Information Processing Systems},
  volume={34},
  pages={6478--6490},
  year={2021}
}

@article{dwork2022,
 author = {Dwork, Cynthia},
 year = {2022},
 title = {Fairness, randomness, and the crystal ball},
 url = {https://www.youtube.com/watch?v=n4XftI9G0fA},
 urldate = {2023-04-08},
 journal = {Munich AI Lectures}
}

@article{gebru2021,
  title={Datasheets for datasets},
  author={Gebru, Timnit and Morgenstern, Jamie and Vecchione, Briana and Vaughan, Jennifer Wortman and Wallach, Hanna and III, Hal Daum{\'e} and Crawford, Kate},
  journal={Communications of the ACM},
  volume={64},
  number={12},
  pages={86--92},
  year={2021},
  publisher={ACM New York, NY, USA}
}

@book{hacking1990,
  title={The Taming of Chance},
  author={Hacking, Ian},
  year={1990},
  publisher={Cambridge University Press}
}

@article{hardt2016,
  title={Equality of opportunity in supervised learning},
  author={Hardt, Moritz and Price, Eric and Srebro, Nati},
  journal={Advances in Neural Information Processing Systems},
  volume={29},
  year={2016}
}

@article{hu2023,
 author = {Hu, Lily},
 year = {2023},
 title = {Causal inference and the problem of variable choice},
 url = {https://www.youtube.com/watch?v=zSwAVwypMSs},
 urldate = {2024-01-03},
 journal = {Social Foundations for Statistics and Machine Learning Workshop}
}

@inproceedings{hutchinson2021,
  title={Towards accountability for machine learning datasets: Practices from software engineering and infrastructure},
  author={Hutchinson, Ben and Smart, Andrew and Hanna, Alex and Denton, Emily and Greer, Christina and Kjartansson, Oddur and Barnes, Parker and Mitchell, Margaret},
  booktitle={Proceedings of the 2021 ACM Conference on Fairness, Accountability, and Transparency},
  pages={560--575},
  year={2021}
}

@article{leonelli2019,
	author = {Leonelli, Sabina},
	journal = {Harvard Data Science Review},
	number = {1},
	year = {2019},
	publisher = {The MIT Press},
	title = {Data governance is key to interpretation: Reconceptualizing data in data science}
}

@article{ludwig2016,
  title={Ontological choices and the value-free ideal},
  author={Ludwig, David},
  journal={Erkenntnis},
  volume={81},
  number={6},
  pages={1253--1272},
  year={2016},
  publisher={Springer}
}

@inproceedings{marx2020,
  title={Predictive multiplicity in classification},
  author={Marx, Charles and Calmon, Flavio and Ustun, Berk},
  booktitle={International Conference on Machine Learning},
  pages={6765--6774},
  year={2020},
  organization={PMLR}
}

@inproceedings{menon2018,
  title={The cost of fairness in binary classification},
  author={Menon, Aditya Krishna and Williamson, Robert C},
  booktitle={Conference on Fairness, Accountability and Transparency},
  pages={107--118},
  year={2018},
  organization={PMLR}
}

@InProceedings{perdomo2024,
  title = 	 {The relative value of prediction in algorithmic decision making},
  author =       {Perdomo, Juan Carlos},
  booktitle = 	 {International Conference on Machine Learning},
  pages = 	 {40439--40460},
  year = 	 {2024},
}

@book{porter1986,
  title={The Rise of Statistical Thinking, 1820-1900},
  author={Porter, Theodore M},
  year={1986},
  publisher={Princeton University Press}
}

@book{porter1995,
  title={Trust in Numbers},
  author={Porter, Theodore M},
  year={1995},
  publisher={Princeton University Press}
}

@inproceedings{roth2023,
  title={Reconciling individual probability forecasts},
  author={Roth, Aaron and Tolbert, Alexander and Weinstein, Scott},
  booktitle={Proceedings of the 2023 ACM Conference on Fairness, Accountability, and Transparency},
  pages={101--110},
  year={2023}
}

@article{rudner1953,
  title={The scientist qua scientist makes value judgments},
  author={Rudner, Richard},
  journal={Philosophy of Science},
  volume={20},
  number={1},
  pages={1--6},
  year={1953},
  publisher={Cambridge University Press}
}

@inproceedings{sambasivan2021,
  title={``{Everyone} wants to do the model work, not the data work'': Data Cascades in High-Stakes AI},
  author={Sambasivan, Nithya and Kapania, Shivani and Highfill, Hannah and Akrong, Diana and Paritosh, Praveen and Aroyo, Lora M},
  booktitle={Proceedings of the 2021 CHI Conference on Human Factors in Computing Systems},
  pages={1--15},
  year={2021}
}

@InProceedings{shirali2024,
  title = 	 {Allocation requires prediction only if inequality is low},
  author =       {Shirali, Ali and Abebe, Rediet and Hardt, Moritz},
  booktitle = 	 {International Conference on Machine Learning},
  pages = 	 {45114--45153},
  year = 	 {2024}
}

@article{strevens2006,
  title={Probability and chance},
  author={Strevens, Michael},
  journal={Encyclopedia of Philosophy, second edition. Macmillan Reference USA, Detroit},
  year={2006}
}

@book{vapnik1982,
  title={Estimation of Dependences Based on Empirical Data},
  author={Vapnik, Vladimir},
  year={1982},
  publisher={Springer}
}

@misc{wang2022,
	type = {{SSRN} {Scholarly} {Paper}},
	title = {Against {Predictive} {Optimization}: {On} the {Legitimacy} of {Decision}-{Making} {Algorithms} that {Optimize} {Predictive} {Accuracy}},
	url = {https://papers.ssrn.com/abstract=4238015},
	author = {Wang, Angelina and Kapoor, Sayash and Barocas, Solon and Narayanan, Arvind},
	year = {2022},
}

@article{watson2023,
  title={Predictive multiplicity in probabilistic classification},
  author={Watson-Daniels, Jamelle and Parkes, David C and Ustun, Berk},
  journal={Proceedings of the AAAI Conference on Artificial Intelligence},
  volume={37},
  number={9},
  pages={10306--10314},
  year={2023}
}

@article{alt1972,
author = {Alt, Franz L.},
title = {Archaelogy of computers: Reminiscences, 1945-1947},
year = {1972},
publisher = {Association for Computing Machinery},
address = {New York, NY, USA},
volume = {15},
number = {7},
journal = {Communications of the ACM},
pages = {693--694}
}

@book{goodfellow2016,
    title={Deep Learning},
    author={Ian Goodfellow and Yoshua Bengio and Aaron Courville},
    publisher={MIT Press},
    note={\url{http://www.deeplearningbook.org}},
    year={2016}
}

@article{vela2022,
  title={Temporal quality degradation in {AI} models},
  author={Vela, Daniel and Sharp, Andrew and Zhang, Richard and Nguyen, Trang and Hoang, An and Pianykh, Oleg S},
  journal={Scientific Reports},
  volume={12},
  number={1},
  pages={11654},
  year={2022},
  publisher={Nature Publishing Group}
}

@article{douglas2000,
  title={Inductive risk and values in science},
  author={Douglas, Heather},
  journal={Philosophy of Science},
  volume={67},
  number={4},
  pages={559--579},
  year={2000},
  publisher={Cambridge University Press}
}

@article{gopalan2023,
  title={Swap agnostic learning, or characterizing omniprediction via multicalibration},
  author={Gopalan, Parikshit and Kim, Michael and Reingold, Omer},
  journal={Advances in Neural Information Processing Systems},
  volume={36},
  pages={39936--39956},
  year={2023}
}

@inproceedings{perdomo2025,
  title={Revisiting the predictability of performative, social events},
  author={Perdomo, Juan Carlos},
  booktitle={International Conference on Machine Learning},
  year={2025}
}

@book{reichenbach1949,
    author = {Reichenbach, Hans},
    publisher = {University of California Press},
    title = {The Theory of Probability},
    year = {1949}
}

@article{roth2025,
  title={Resolving the reference class problem at scale},
  author={Roth, Aaron and Tolbert, Alexander},
  journal={Philosophy of Science},
  volume={92},
  number={4},
  pages={868--882},
  year={2025}
}

@inproceedings{thorp1998,
  title={The invention of the first wearable computer},
  author={Thorp, Edward O},
  booktitle={Digest of Papers. Second International Symposium on Wearable Computers},
  pages={4--8},
  year={1998},
  organization={IEEE}
}

@book{durr2009,
  title={Bohmian Mechanics: The Physics and Mathematics of Quantum Theory},
  author={Detlef and D{\"u}rr, Detlef and Teufel, Stefan},
  year={2009},
  publisher={Springer}
}

@inproceedings{cooper2024,
  title={Arbitrariness and social prediction: The confounding role of variance in fair classification},
  author={Cooper, A Feder and Lee, Katherine and Choksi, Madiha Zahrah and Barocas, Solon and De Sa, Christopher and Grimmelmann, James and Kleinberg, Jon and Sen, Siddhartha and Zhang, Baobao},
  booktitle={Proceedings of the AAAI Conference on Artificial Intelligence},
  volume={38},
  number={20},
  pages={22004--22012},
  year={2024}
}

@book{gigerenzer2008,
  title={Rationality for mortals: How people cope with uncertainty},
  author={Gigerenzer, Gerd},
  year={2008},
  publisher={Oxford University Press}
}

@article{jaynes1985,
  title={Some random observations},
  author={Jaynes, Edwin T},
  journal={Synthese},
  volume={63},
  pages={115--138},
  year={1985},
  publisher={Springer}
}

@article{kingma2013,
  title={Auto-encoding variational Bayes},
  author={Kingma, Diederik P and Welling, Max},
  journal={arXiv preprint arXiv:1312.6114},
  year={2013}
}

@article{meng2022,
  title={Comments on "Statistical inference with non-probability survey samples"-Miniaturizing data defect correlation: A versatile strategy for handling non-probability samples},
  author={Meng, Xiao-Li},
  journal={Survey Methodology},
  volume={48},
  number={2},
  pages={339--360},
  year={2022}
}

@article{shafer2008,
  title={A tutorial on conformal prediction},
  author={Shafer, Glenn and Vovk, Vladimir},
  journal={Journal of Machine Learning Research},
  volume={9},
  number={3},
  year={2008}
}

@article{spiegelhalter2024,
  title={Does probability exist?},
  author={Spiegelhalter, David},
  journal={Nature},
  volume={636},
  pages={19},
  year={2024}
}

@article{cartwright1999,
  title={The limits of exact science, from economics to physics},
  author={Cartwright, Nancy},
  journal={Perspectives on Science},
  volume={7},
  number={3},
  pages={318--336},
  year={1999},
  publisher={MIT Press}
}

@inproceedings{hebert2018,
  title={Multicalibration: Calibration for the (computationally-identifiable) masses},
  author={H{\'e}bert-Johnson, Ursula and Kim, Michael and Reingold, Omer and Rothblum, Guy},
  booktitle={International Conference on Machine Learning},
  pages={1939--1948},
  year={2018},
  organization={PMLR}
}

@inproceedings{jacobs2021,
  title={Measurement and fairness},
  author={Jacobs, Abigail Z and Wallach, Hanna},
  booktitle={Proceedings of the 2021 ACM Conference on Fairness, Accountability, and Transparency},
  pages={375--385},
  year={2021}
}

@inproceedings{perdomo2025wisconsin,
  title={Difficult lessons on social prediction from {Wisconsin} public schools},
  author={Perdomo, Juan Carlos and Britton, Tolani and Hardt, Moritz and Abebe, Rediet},
  booktitle={Proceedings of the 2025 ACM Conference on Fairness, Accountability, and Transparency},
  pages={2682--2704},
  year={2025}
}

@article{strikwerda2021,
  title={Predictive policing: The risks associated with risk assessment},
  author={Strikwerda, Litska},
  journal={The Police Journal},
  volume={94},
  number={3},
  pages={422--436},
  year={2021},
  publisher={SAGE Publications Sage UK: London, England}
}

@book{potochnik2017,
  title={Idealization and the Aims of Science},
  author={Potochnik, Angela},
  year={2017},
  publisher={University of Chicago Press}
}

@book{appiah2017,
  title={As If: Idealization and Ideals},
  author={Appiah, Anthony},
  year={2017},
  publisher={Harvard University Press}
}

@article{heljakka2022,
  title={Disentangling model multiplicity in deep learning},
  author={Heljakka, Ari and Trapp, Martin and Kannala, Juho and Solin, Arno},
  journal={arXiv preprint arXiv:2206.08890},
  year={2022}
}

@inproceedings{mitchell2019,
  title={Model cards for model reporting},
  author={Mitchell, Margaret and Wu, Simone and Zaldivar, Andrew and Barnes, Parker and Vasserman, Lucy and Hutchinson, Ben and Spitzer, Elena and Raji, Inioluwa Deborah and Gebru, Timnit},
  booktitle={Proceedings of the 2019 Conference on Fairness, Accountability, and Transparency},
  pages={220--229},
  year={2019}
}

@article{vaswani2017,
  title={Attention is all you need},
  author={Vaswani, Ashish and Shazeer, Noam and Parmar, Niki and Uszkoreit, Jakob and Jones, Llion and Gomez, Aidan N and Kaiser, {\L}ukasz and Polosukhin, Illia},
  journal={Advances in Neural Information Processing Systems},
  volume={30},
  year={2017}
}

@article{dwork2025,
  title={From fairness to infinity: Outcome-indistinguishable (omni) prediction in evolving graphs},
  author={Dwork, Cynthia and Hays, Chris and Immorlica, Nicole and Perdomo, Juan C and Tankala, Pranay},
  journal={The Thirty Eighth Annual Conference on Learning Theory},
  year={2025}
}

@article{holtgen2024,
  title={Practical foundations for probability: Prediction methods and calibration},
  author={H{\"o}ltgen, Benedikt},
  year={2024},
  journal={PhilPapers preprint},
  url={https://philpapers.org/rec/HLTPFF}
}

@article{holtgen2025ftu,
  title={Reconsidering Fairness Through Unawareness from the Perspective of Model Multiplicity},
  author={H{\"o}ltgen, Benedikt and Oliver, Nuria},
  journal={EAAMO '25: Proceedings of the 5th ACM Conference on Equity and Access in Algorithms, Mechanisms, and Optimization},
  year={2025}
}

@inproceedings{fletcher2025credal,
  title={CREDAL: Close Reading of Data Models},
  author={Fletcher, George and Nahurna, Olha and Prytula, Matvii and Stoyanovich, Julia},
  booktitle={Proceedings of the Workshop on Human-In-the-Loop Data Analytics},
  pages={1--7},
  year={2025}
}

@book{joque2022revolutionary,
  title={Revolutionary mathematics: Artificial intelligence, statistics and the logic of capitalism},
  author={Joque, Justin},
  year={2022},
  publisher={Verso Books}
}

@book{beuving2015,
  title={Doing qualitative research: The craft of naturalistic inquiry},
  author={Beuving, Joost and Vries, Geert},
  year={2015},
  publisher={Amsterdam University Press}
}

\appendix


\section{Proofs}

We consider a population of $k+l$ data points of which $l$ are taken for the training set.
$v_{tr}(\alpha)$ and $u(\alpha)$ are the error ratios of model $\alpha$ on the train set and the whole urn, respectively.
We consider a theoretical test set with $l$ samples and error rate $v_{te}(\alpha)$.
We omit `$(\alpha)$' when the model is clear or irrelevant.

We define $u'$ to be the error rate in the remaining urn after taking the train set, \emph{i.e.}
\begin{equation}\label{eq:uprime}
    u' = \frac{l+k}{k} u - \frac{l}{k} v_{tr} = u + \frac{l}{k}(u-v_{tr})
\end{equation}
since $u(k+l) = v_{tr}l + u'k$.

For the proof of the theorem, we need modified, finite-urn versions of Vapnik's Symmetrisation Lemma.

\begin{replemma}{lem:symm1}[Symmetrisation Lemma, version 1]
    \ \\
    Take an urn with $l+k$ balls, with $k \geq l > 2/\epsilon$.
    Let $u(\alpha)$ denote the error rate of $\alpha$ on the whole urn.
    Then
        \begin{align*}
            \mu &\left[ \sup_\alpha \left| u(\alpha) - v_{tr}(\alpha) \right|  > \epsilon \right]\\
            &\quad\quad< 2 \cdot \mu \left[ \sup_\alpha \left| v_{te}(\alpha) - v_{tr}(\alpha) \right| > \left(\frac{1}{2}+\frac{l}{k}\right) \epsilon \right].
        \end{align*}
\end{replemma}

\emph{Proof.}
Drawing $2l$ balls without replacement and partitioning them into train and test set is the same as first taking $l$ for the train set without replacement and then taking $l$ for the test set, since we are only concerned with numbers of combinations.

The proof is based on Vapnik's but improves on a na\"ive application by taking into account that the test set is sampled from a changed urn with a slightly different error ratio.

We give a lower bound for the RHS by showing that draws that break the LHS bound have at least a $1/2$-probability that they also break the RHS bound.
Put differently, of all combinations that break the LHS bound, at least 1/2 also break the RHS bound.

Consider a draw for which there is an $\alpha^*$ with $u(\alpha^*) - v_{tr}(\alpha^*) > \epsilon$ (analogous for $v_{tr}(\alpha^*) - u(\alpha^*) > \epsilon$).

So assuming that $u - v_{tr} > \epsilon$ we see that $u'- v_{te} < \epsilon/2$ would guarantee
\begin{align}
   v_{te} - v_{tr}
   &= (v_{te}-u') + (u'-u) + (u-v_{tr}) \\
   &> - \epsilon/2 + (u'-u) + \epsilon \\
   &= \epsilon/2  + (u'-u)
\end{align}
and thus, since 
\begin{equation}
    u'-u = \left(u + \frac{l}{k} (u-v_{tr})\right) - u = \frac{l}{k} (u-v_{tr}) > \frac{l}{k} \epsilon
\end{equation}
we get
\begin{equation}
   |v_{te} - v_{tr}| > \left(\frac{1}{2} + \frac{l}{k}\right) \epsilon. 
\end{equation}

This means that
\begin{align}
    \mu(u' - v_{te} < \epsilon/2 | u - v_{tr} > \epsilon) \cdot
    \mu \left[ \sup_\alpha \left| u(\alpha) - v_{tr}(\alpha) \right|  > \epsilon \right]& 
    \label{eq:conditional}
    < \mu \left[ \sup_\alpha \left| v_{te}(\alpha) - v_{tr}(\alpha) \right| > \left(\frac{1}{2}+\frac{l}{k}\right) \epsilon \right].
\end{align}

Now $\mu(u' - v_{te} < \epsilon/2 | u - v_{tr} > \epsilon)$ is the probability that
for an urn with $k$ balls and $ku' > ku + l \epsilon$ red ones, we draw $l$ balls of which at least $(u' - \epsilon/2) \cdot l$ are red.
Now 
\begin{equation}\label{eq:half}
    \mu(u' - v_{te} < \epsilon/2 | u - v_{tr} > \epsilon) > 1/2,
\end{equation}
given that the expected number of balls drawn is $u' l$ and $\epsilon l / 2 > 1$ guarantees that there is an integer between $u' l$ and $(u' + \epsilon/2) \cdot l$.
Now (\ref{eq:conditional}) and (\ref{eq:half}) taken together prove the lemma. 
\qed

\begin{replemma}{lem:symm2}[Symmetrisation Lemma, version 2]
    \ \\
    Take an urn with $l+k$ balls, with $k \geq l > 2/\epsilon$.
    Let $u'(\alpha)$ denote the error rate of $\alpha$ on the remaining urn.
    Then
        \begin{align*}
            \mu \left[ \sup_\alpha \left| u'(\alpha) - v_{tr}(\alpha) \right|  > \epsilon \right]
            < 2 \cdot \mu \left[ \sup_\alpha \left| v_{te}(\alpha) - v_{tr}(\alpha) \right| > \epsilon / 2 \right].
        \end{align*}
\end{replemma}

\textit{Proof.}
Same proof as in Vapnik' Symmetrisation Lemma but with $u'(\alpha)$ (see (\ref{eq:uprime})) instead of $P(\alpha)$:

We show that of all draws where some $\alpha$ breaks the LHS bound, at least half also break the RHS bound.

Consider draws that break the LHS and fix an $\alpha^*$ for which the LHS bound is broken, and assume w.l.o.g. $u' - v_{tr} > \epsilon$ (the reverse case is analogous).

Then if $u'-v_{te} < \epsilon/2$, we get
\begin{equation}
    v_{te} - v_{tr} = (v_{te} - u') - (u' - v_{tr}) > \epsilon - \epsilon/2 = \epsilon/2.
\end{equation}

As the reverse case is analogous, we can bound
\begin{align}
    \mu(u' - v_{te} < \epsilon/2 | u' - v_{tr} > \epsilon) \cdot
    \mu \left[ \sup_\alpha \left| u'(\alpha) - v_{tr}(\alpha) \right|  > \epsilon \right]
    <\ \mu \left[ \sup_\alpha \left| v_{te}(\alpha) - v_{tr}(\alpha) \right| > \epsilon/2 \right]&. \label{eq:conditional_uprime}
\end{align}

Now $\mu(u' - v_{te} < \epsilon/2 | u - v_{tr} > \epsilon)$ is the probability that
for an urn with $k$ balls and $ku' > ku + l \epsilon$ red ones, we draw $l$ balls of which at least $(u' - \epsilon/2) \cdot l$ are red.
Now 
\begin{equation}\label{eq:half_uprime}
    \mu(u' - v_{te} < \epsilon/2 | u - v_{tr} > \epsilon) > 1/2,
\end{equation}
given that the expected number of balls drawn is $u' l$ and $\epsilon l / 2 > 1$ guarantees that there is an integer between $u' l$ and $(u' + \epsilon/2) \cdot l$.

Now (\ref{eq:conditional_uprime}) and (\ref{eq:half_uprime}) taken together prove the lemma. 
\qed

\begin{reptheorem}{LT_theorem}
    \ \\
    Take a set\footnote{Strictly speaking, these are multi-sets, but we conform with common lingo here.} of $l+k$ data points in $\mathcal{X} \times \{0,1\}$ and draw a training set of size $l$ without replacement.
    Consider a hypothesis class $\Lambda$ of finite VC dimension $h$.
    Denote by $u(\alpha)$, $v_{tr}(\alpha)$, and $u'(\alpha)$ the error rate of $f_\alpha \in \Lambda$ on the whole set of $k+l$ points, on the training set of $l$ points, and on the remaining $k$ points, respectively.

    Then for $k \geq l > 2/\epsilon$ with $\epsilon >0$, we can bound the proportion of draws in which the generalisation error exceeds $\epsilon$ by
    \begin{align*} 
    \mu \left[ \sup_{\alpha \in \Lambda} \left| u(\alpha) - v_{tr}(\alpha) \right| > \epsilon \right]
    \ < \
    9 \frac{(2l)^h}{h!} \cdot \exp \left(-\epsilon^2l \left(\frac{1}{2}+\frac{l}{k}\right)^2\right) 
    \end{align*}
    and 
    \begin{align*} 
    \mu \left[ \sup_{\alpha \in \Lambda}\left| u'(\alpha) - v_{tr}(\alpha) \right| > \epsilon \right]
    \ < \ 
    9 \frac{(2l)^h}{h!} \cdot \exp \left(-\epsilon^2l/4\right),\quad\quad\quad\
    \end{align*} 
    where $\mu$ is the counting measure on possible draws of size $l$.
\end{reptheorem}

\textit{Proof.}
It is sufficient to bound
\begin{equation}
    \mu \left[ \sup_\alpha \left| v_{te}(\alpha) - v_{tr}(\alpha) \right| > \epsilon \right] < m^S(2l) \cdot 3 \exp\left(-\epsilon^2l\right)
\end{equation}
for two sets of size $l$ the same way as in Vapnik's proof of Theorem A.2 and use the respective Symmetrisation Lemma.
The growth function $m^s(l)$ is bounded by $m^S(l) < 1.5 \frac{l^h}{h!}$, see Theorem 6.6 in \citep[p. 154]{vapnik1982}.
\qed
\ \\

Note that both versions converge to Vapnik's result for $k \to \infty$, as this means $l/k \to 0$ and $u' \to u$, respectively.
Also note that our second version is stronger (the bound is tighter) than the original version in the sense that the LHS is larger while the RHS is the same.

\section{Experiments}

The code for all experiments is available at \url{https://github.com/ben-hoeltgen/gen-d-illustrations}.

\subsection{Additional experiments on model degradation}
\label{app:degradation}

\begin{figure}[h]
    \centering
    \includegraphics[width=\linewidth]{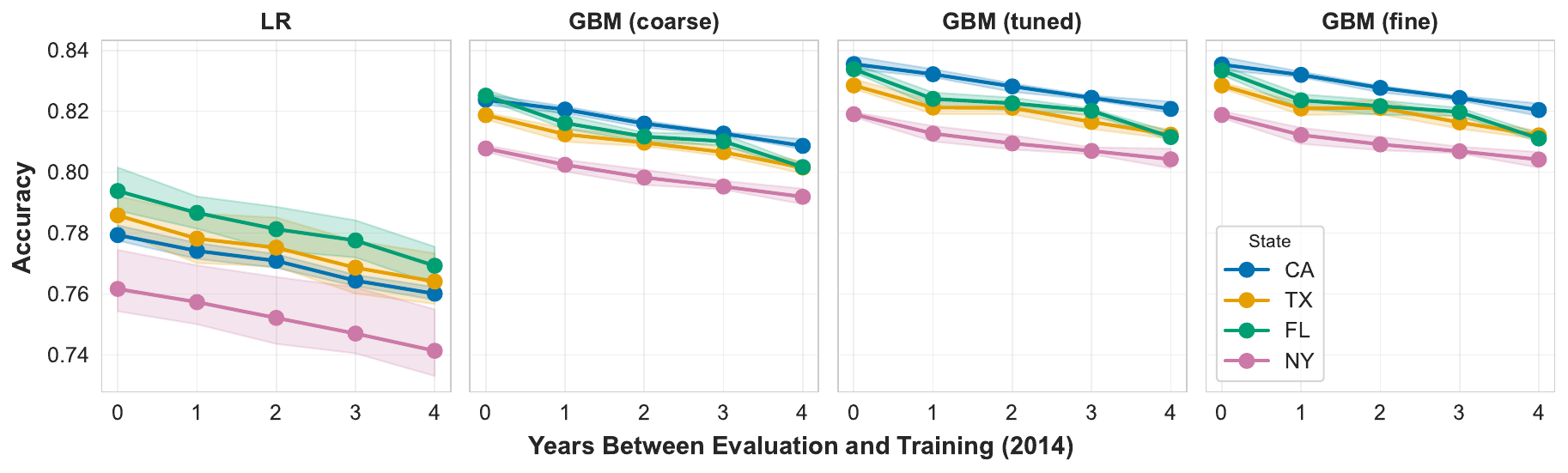}
    \caption{Temporal performance degradation on \texttt{ACS Income} for various models and different US states: Models were trained on 2014 data and tested on subsequent years. Coloured shadows visualise min and max over 10 seeds.}
    \label{fig:degradation}
\end{figure}

We ran additional experiments where models were trained on a random subset of the 2014 data of the four biggest states (10 subsets for 10 models) and tested on subsequent years.
We observe performance degradation for each subsequent year (Figure~\ref{fig:degradation}), comparable to that reported on the whole US data by \citet{ding2021}.

\subsection{More details on the experimental setup}
\label{app:exp_details}

For the results reported in Figures~\ref{fig:multiplicity} and~\ref{fig:degradation}, we used \texttt{scikit-learn} default parameters for \textit{LR} except setting \texttt{max-iter}=1000.
We used default parameters of \texttt{scikit-learn} for \textit{GBM (coarse)}; for \textit{GBM (fine)}, we took the parameters that we found to be optimal for the whole US data of \texttt{ACS Income} in previous work \citep{holtgen2025ftu}, very similar to those of \citet{cruz2024}; we tuned them to try to improve performance while making it coarser, resulting in \textit{GBM (tuned)}.
Details on the GBM configurations are given in Table~\ref{tab:gbm_hyperparameters}.
Each seed uses 50\% of the respective year for training.
For the model pairs that are compared in Figure~\ref{fig:multiplicity}, all data on which the 2017 model is trained is also used to train its respective 2014-2017 partner model.

\begin{table}[h]
    \caption{Hyperparameter configurations for the used GBM settings; `GBM (coarse)' is the \texttt{scikit-learn} default.}
    \label{tab:gbm_hyperparameters}
    \small
    \begin{tabular}{lccc}
        \toprule
        \textbf{Hyperparameter} & \textbf{GBM (coarse)} & \textbf{GBM (tuned)} & \textbf{GBM (fine)} \\
        \midrule
        n\_estimators & 100 & 400 & 500 \\
        max\_depth & 3 & 8 & 10 \\
        max\_leaf\_nodes & None & 50 & 50 \\
        min\_samples\_leaf & 1 & 500 & 500 \\
        \bottomrule
    \end{tabular}
\end{table}

\end{document}